%% file: bare_conf_compsoc.tex
\documentclass[conference,compsoc]{IEEEtran}
\ifCLASSOPTIONcompsoc
  \usepackage[nocompress]{cite}
\else
  \usepackage{cite}
\fi
\ifCLASSINFOpdf
\else
\fi
\usepackage{amsmath}
\usepackage{booktabs}
\usepackage{graphicx}
\usepackage{pdflscape} 
\usepackage{multirow} 
\usepackage{caption}
\usepackage{titlesec}
\usepackage{algorithm} 
\usepackage{algpseudocode} 
\usepackage[table,xcdraw]{xcolor}
\usepackage{colortbl}
\usepackage{adjustbox} 
\usepackage{hyperref}

\titleformat{\subsubsection}
  {\normalfont\normalsize\bfseries}{\thesubsubsection}{1em}{}
\titlespacing*{\subsubsection}
  {0pt}{3.25ex plus 1ex minus .2ex}{1.5ex plus .2ex}

\IEEEoverridecommandlockouts
\begin{document}

%
\title{OT-Attack: Enhancing Adversarial Transferability of
Vision-Language Models via Optimal Transport Optimization}



%
\author{\IEEEauthorblockN{Dongchen Han \textsuperscript{\rm 1, $\dag$},
Xiaojun Jia \textsuperscript{\rm 2, $\dag$ \thanks{$^{\dag}$ The first two authors contributed equally to this work (me@handongchen.com \& jiaxiaojunqaq@gmail.com).} },
Yang Bai \textsuperscript{\rm 3}, 
Jindong Gu \textsuperscript{\rm 4},
Yang Liu \textsuperscript{\rm 2} and
Xiaochun Cao \textsuperscript{\rm 1, $^{\star}$ \thanks{$^{\star}$ Corresponding author: caoxiaochun@mail.sysu.edu.cn}}
}

\IEEEauthorblockA{\textsuperscript{\rm 1} Shenzhen Campus of Sun Yat-sen University}

\IEEEauthorblockA{\textsuperscript{\rm 2} Nanyang Technological University}

\IEEEauthorblockA{\textsuperscript{\rm 3} Tsinghua University}

\IEEEauthorblockA{\textsuperscript{\rm 4} University of Oxford}
}


\maketitle

\input{section/Abstract}


%
\IEEEpeerreviewmaketitle

\input{section/Introduction}

\input{section/Related_work}
\input{section/Approach}

\input{section/Experiments}

\input{section/Conclusion}

\bibliographystyle{IEEEtran}
\bibliography{refer}

\end{document}

%% file: section/Abstract.tex
\begin{abstract}


Vision-language pre-training (VLP) models demonstrate impressive abilities in processing both images and text. However, they are vulnerable to multi-modal adversarial examples (AEs). Investigating the generation of high-transferability adversarial examples is crucial for uncovering VLP models' vulnerabilities in practical scenarios.
Recent works have indicated that leveraging data augmentation and image-text modal interactions can enhance the transferability of adversarial examples for VLP models significantly. However, they do not consider the optimal alignment problem between data-augmented image-text pairs. This oversight leads to adversarial examples that are overly tailored to the source model, thus limiting improvements in transferability.
In our research, we first explore the interplay between image sets produced through data augmentation and their corresponding text sets. We find that augmented image samples can align optimally with certain texts while exhibiting less relevance to others. Motivated by this, we propose an Optimal Transport-based Adversarial Attack, \emph{dubbed} OT-Attack. The proposed method formulates the features of image and text sets as two distinct distributions and employs optimal transport theory to determine the most efficient mapping between them. This optimal mapping informs our generation of adversarial examples to effectively counteract the overfitting issues. Extensive experiments across various network architectures and datasets in image-text matching tasks reveal that our OT-Attack outperforms existing state-of-the-art methods in terms of adversarial transferability.
\end{abstract}



%% file: section/Introduction.tex
\section{Introduction}

\begin{figure}
    \centering
    \includegraphics[width=1\columnwidth]{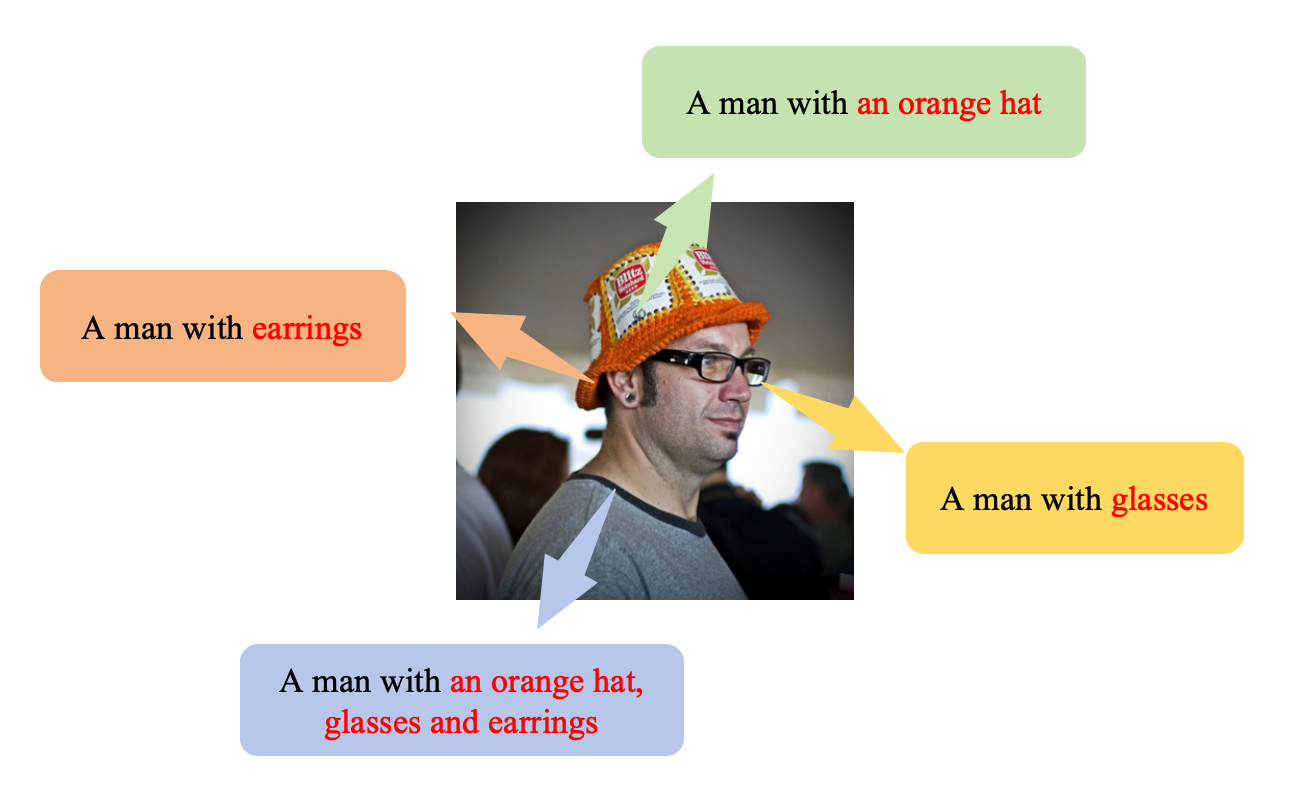}
    \caption{An image example, after undergoing various data augmentation strategies, tends to focus on different image contents. Consequently, it can better align with specific text content while maintaining limited relevance to others. Thus, using a uniform standard to assess this relationship is unsuitable, which highlights the limitation of the existing state-of-the-art SGA method.}
    \label{fig:home}
\end{figure}

Vision-Language Pre-trained (VLP) models have shown outstanding performance in various downstream tasks, including image-text matching, image captioning, visual question answering, and visual grounding. Despite their impressive capabilities, these models encounter significant security challenges in real-world applications~\cite{lei2021understanding,zhou2020unified,bao2022vlmo,hu2022scaling}. Most of the prior researches~\cite{jia2022adversarial,zhang2022towards,yin2023vlattack,zhou2023advclip,jia2023improving} focused on white-box attacks, where there is complete access to the model's structure, weights, and gradients. But they are not applicable to black-box models, such as GPT-4~\cite{DBLP:journals/corr/abs-2303-08774}. 


\par Fortunately, existing works have demonstrated that adversarial examples perturbed on white-box models remain effective on certain black-box models~\cite{goodfellow2014explaining,papernot2016transferability}. It indicates that adversarial examples generated via a proxy model can still mislead the judgment of black-box models due to their transferability~\cite{xie2019improving,lin2019nesterov,dong2019evading,jia2020adv,long2022frequency,jia2022prior}. The most ideal attack scenario, in reality, is one where adversarial examples remain effective even in the absence of detailed knowledge about the model's inner workings, such as its model architecture, weights, and gradients, etc~\cite{han2023interpreting,gu2023survey}. 
Motivated by the practical significance of black-box adversarial attacks and adversarial transferability~\cite{gubri2022lgv,qin2022boosting,byun2022improving,waseda2023closer}, in this paper, we primarily study the transferability of adversarial examples within Vision-Language Pre-trained (VLP) models.



Recent research in the field of VLP models has revealed that single-modality adversarial attacks, such as employing BERT-Attack\cite{li2020bert} on text or PGD attack\cite{madry2017towards} on images, yield limited attack success rates, even against white-box models. This insight led to the exploration of more integrated approaches. For instance, the Sep-Attack method, which simultaneously uses BERT-Attack on text and PGD on images without considering interactions between these modalities, has shown some effectiveness but lacks in fully exploiting the potential synergies of multi-modal attacks. Building upon these findings, Zhang \emph{et al.} proposed the Co-Attack\cite{zhang2022towards} method, which goes a step further by fusing information between modalities using individual image-text pairs. This approach underscored the importance of intermodal guidance for improving the transferability of adversarial examples in VLP models. Further advancing this field, Lu \emph{et al.} introduced the Set-level Guidance Attack (SGA) \cite{lu2023set}, which expands upon the concept of individual image-text pairings to a set-level alignment. The SGA methodology employs data augmentation to diversify the image set and pairs these images with multiple textual descriptions. This comprehensive approach, which integrates intermodal guidance, not only addresses the limitations observed in Sep-Attack but also achieves state-of-the-art results, highlighting the effectiveness in enhancing the transferability of adversarial examples against VLP models.

\begin{figure}
    \centering
    \includegraphics[width=1\columnwidth]{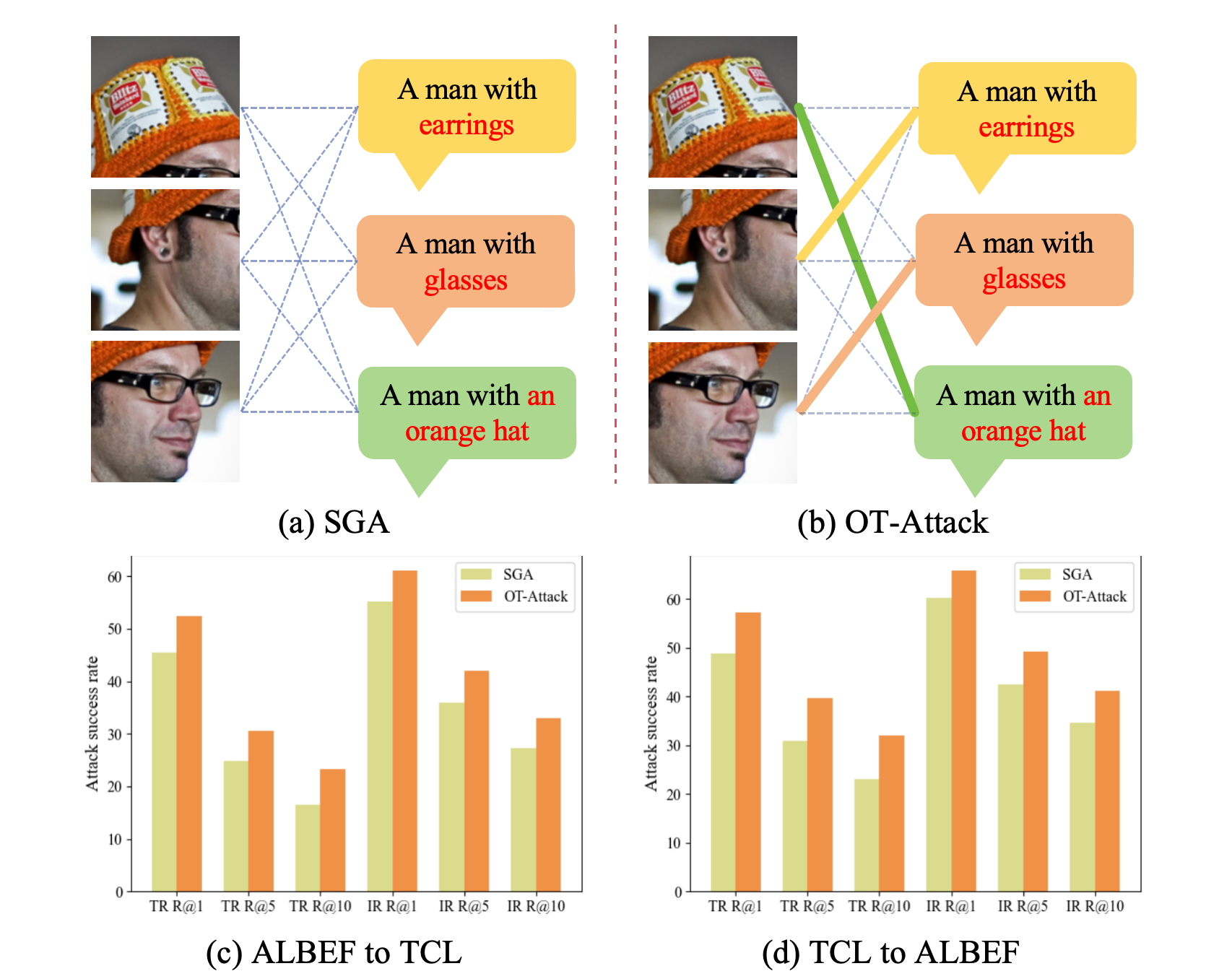}
    \caption{Comparative analysis of Set-level Guidance Attack (SGA) methods and their ITR attack success rates. Panel \textbf{(a)} illustrates the conventional SGA approach where image and text sets are averaged to establish pair-wise matches. Panel \textbf{(b)} showcases our proposed method, OT-Attack, where images are matched to texts based on optimal transport theory to enhance matching accuracy. Panels \textbf{(c)} and \textbf{(d)} depict the attack success rates for our method OT-Attack versus traditional SGA, with ALBEF and TCL models serving alternately as the source and target. \textbf{The bar charts indicate that our adversarial examples outperform SGA across all metrics}, demonstrating superior effectiveness in disrupting ITR performance.}
    \label{fig:home2}
\end{figure}

However, SGA has limitations in improving the transferability of adversarial examples for vision-language pre-training models. Specifically, it aligns the features of images and texts at a collective level but does not adequately address the optimal matching of post-augmentation image examples with their corresponding texts. For example, when images undergo data augmentation, such as zooming, which enlarges certain parts of an image while omitting others, the resulting examples may not align equally well with their textual descriptions. SGA calculates its loss by averaging similarities across all pairs, which means that even if there is an optimal match for the augmented data, its influence on the final loss is obviously reduced. This process can undermine the benefits of data augmentation and modality interactions in enhancing adversarial transferability.

In this paper, we address this issue by incorporating the theory of optimal transport~\cite{villani2009optimal}. We treat the feature sets of augmented images and texts as two distinct distributions and aim to establish the optimal transport scheme between them. The distinction between our method and SGA, along with a comparative overview of the results, is depicted in Figure \ref{fig:home2}.
In detail, we integrate optimal transport theory to analyze data-augmented image sets and text sets as distinct distributions. This holistic consideration allows us to incorporate similarity into the cost matrix and calculate the optimal transport scheme. Consequently, we compute the total transfer cost between these distributions, guiding the generation of adversarial examples. Our method achieves a more balanced matching relationship between the augmented image set and the text set. As depicted in Figure \ref{fig:home}, an image example, after undergoing data augmentation, may align more closely with a particular text description. Applying optimal transport theory addresses this by balancing the effects of data augmentation and modality interactions, which in turn reduces overfitting and improves the transferability of adversarial examples.

Experiments conducted on various models including ALBEF~\cite{li2021align}, TCL~\cite{yang2022vision}, and CLIP~\cite{radford2021learning}, and utilizing well-known datasets like Flickr30K~\cite{plummer2015flickr30k} and MSCOCO~\cite{lin2014microsoft}, quantitatively demonstrate the effectiveness of our approach. We focus on image-text and text-image matching as our primary downstream tasks, assessing the adversarial examples' transferability through their attack success rates on black-box models. The results exceed the current state-of-the-art on both datasets. Additionally, to assess cross-task transferability, we conduct experiments on image captioning and visual grouping, which confirm the efficacy of our OT-Attack method in varied downstream tasks. Additionally, we assess widely-used business models such as GPT-4 and Bing Chat under increased perturbation intensity, and we are able to effectively break them. Our research indicates that extra textural perturbations cause confusion in the decision-making processes of these models. The proposed method is positioned as a robust solution for uncovering and leveraging widespread vulnerabilities across the spectrum of VLP models. The key contributions of 
 in this paper are summarized in three aspects:

\begin{enumerate}
    \item Our proposed OT-Attack improves the SGA framework by ensuring a balanced match between image and text sets after data augmentation.
    \item We innovatively utilize Optimal Transport theory in examining adversarial example transferability in VLP models, promoting a more profound and thorough alignment between data-augmented images and textual descriptions.
    \item Extensive experiments establish that our method generates adversarial examples with superior transferability compared to existing state-of-the-art techniques. \textbf{Furthermore, our OT-Attack can successfully break current business models like GPT-4 and Bing Chat.}
\end{enumerate}

%% file: section/Related_work.tex
\section{Background And Related Work}

\subsection{Vision-Language Pre-training Models}

Vision-language pre-training (VLP)~\cite{chen2023vlp} is a pivotal technique in augmenting multimodal task performance, capitalizing on extensive pre-training with image-to-text pairs. Traditionally, much of the research in this area has relied on pre-trained object detectors, using region features to create vision-language representations. However, the advent of Vision Transformer (ViT)~\cite{dosovitskiy2020image,han2022survey} has instigated a methodological shift. Increasingly, studies are advocating for the adoption of ViT in image encoding, which involves an end-to-end process of transforming inputs into patches.VLP models can be broadly classified into two categories: fused and aligned VLP models. Fused VLP models, as exemplified by architectures like ALBEF~\cite{li2021align} and TCL~\cite{yang2022vision}, utilize individual unimodal encoders for processing token and visual feature embeddings. These models then employ a multimodal encoder to amalgamate image and text embeddings, crafting comprehensive multimodal representations. Conversely, aligned VLP models, such as CLIP, use unimodal encoders to independently process image and text modality embeddings.

\subsection{Vision-Language Tasks}

\textbf{Image-text Retrieval.} Image-Text Retrieval (ITR)\cite{cao2022image,li2023commonsense} is a task that retrieves relevant instances from a database using one modality (image or text) to query the other. It splits into image-to-text retrieval (TR) and text-to-image retrieval (IR). Models like ALBEF and TCL calculate semantic similarity scores between image-text pairs for initial ranking, then employ a multimodal encoder for final ranking. Conversely, models like CLIP~\cite{radford2021learning} directly rank based on similarity in an unimodal embedding space, showcasing varied ITR methodologies.

\textbf{Image Captioning.} Image captioning~\cite{hossain2019comprehensive,ghandi2023deep} involves generating textual captions for images and is crucial in Vision-Language Pre-training (VLP) models. This task requires converting visual content into coherent, contextually relevant text, differing from image-text retrieval which is about finding the most relevant content from a database based on the opposite modality queries.

\textbf{Visual Grounding.} Visual Grounding~\cite{deng2018visual,yang2023improving}, also termed referential expression comprehension, is key in both computer vision and NLP. It entails identifying and locating objects or regions in an image as per language descriptions, requiring a precise mapping of text to visual elements.

\subsection{Transferability of Adversarial Examples}

In the field of adversarial attacks, there are two main categories: white-box and black-box approaches. White-box attacks assume full knowledge of the target model's structure and parameters, a scenario often not feasible in real-world black-box settings. In natural language processing (NLP), predominant methods typically involve textual manipulation, such as altering or replacing specific tokens, exemplified by BERT-Attack, introduced by Li \emph{et al.}\cite{li2020bert}. In computer vision, white-box attacks have seen significant developments with methods like the Fast Gradient Sign Method (FGSM) by Goodfellow \emph{et al.}\cite{goodfellow2014explaining}, Carlini \& Wagner (C\&W) by Carlini and Wagner\cite{madry2017towards}, and the Momentum Iterative Method (MIM) by Dong \emph{et al.} \cite{dong2018boosting}. These methods exploit the gradient information of the models to craft effective adversarial examples. Separately, Projected Gradient Descent (PGD), proposed by Madry \emph{et al.}\cite{madry2017towards}, marks a notable advancement in the field. PGD iteratively adjusts images in small steps along the gradient of the loss function, making it a potent tool for creating subtle yet effective adversarial images. This approach has proven particularly effective in attacking the visual components of models.

Building on the foundations laid by PGD and BERT-Attack, Co-Attack, introduced by Xie \emph{et al.}\cite{zhang2022towards}, emerged as a method integrating both visual and textual modalities. This approach exploits the synergies between image and text within VLP models, targeting their multimodal nature. The recent innovation in this domain is the Set-level Guidance Attack (SGA), developed by Zhang \emph{et al.}\cite{lu2023set}. SGA uses data augmentation to generate multiple image sets, aligning them with various text captions. This approach not only complements the multimodal essence of VLP models but also significantly enhances the transferability of adversarial examples across a spectrum of black-box models. The progression from individual-focused PGD and BERT-Attack to the integrated Co-Attack, and finally to the comprehensive SGA, illustrates an evolving landscape of adversarial strategies against VLP models.

\subsection{Optimal Transport}
Optimal Transport (OT), a concept first introduced by Monge~\cite{villani2009optimal}, was initially developed to minimize logistical costs in transporting goods. In modern times, especially within the realms of machine learning and computer vision, OT has become a prominent tool due to its effectiveness in comparing and aligning distributions that are represented as feature sets~\cite{peyre2017computational}. Its unique ability to match distributions has led to its widespread application in various theoretical and practical areas. This includes its use in generative models and structural alignments involving sequences~\cite{arjovsky2017wasserstein}, graphs~\cite{xu2019gromov}, and image matching~\cite{zhang2020deepemd,liu2021multi,zhao2021towards}. The versatility of OT also extends to other distribution-centric tasks like clustering, distribution estimation, and causal discovery. 

%% file: section/Approach.tex
\section{Approach}

\begin{figure*}
    \centering
    \includegraphics[width=1.0\textwidth]{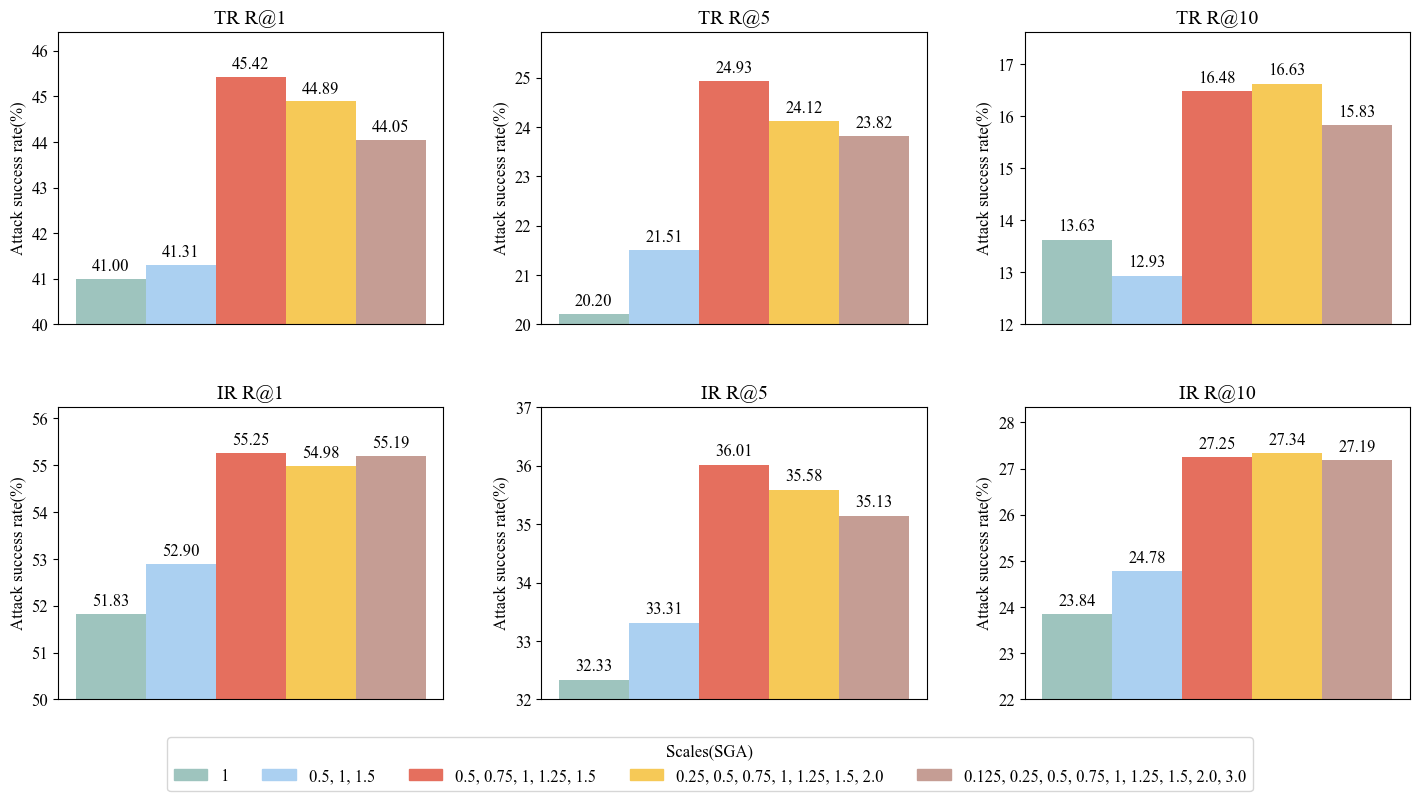}
    \caption{Utilizing the SGA method, this caption presents the attack success rates when the augmented image set, originating from the ALBEF source model and targeting the TCL model, contains 1, 3, and up to 9 images. The overall trend progresses from an increase to a decrease in success rates with the addition of examples, illustrating the effectiveness of the image set and the diminishing performance on the black-box model with an excessive number of examples.}
    \label{fig:overfitting}
\end{figure*}

In this section, we first introduce the treated model in this work, which consists of 
the adversary’s goals and the adversary’s capabilities. Then we present the observation of the previous works and the corresponding analysis. Finally, we introduce our proposed method in detail.

\subsection{Threat Model}
 \textbf{Adversary’s goals.} Let us consider a white-box model 
 \( M_{\text{white}} \) with an encoding function \( f_{\text{white}} \) and an original image \( I_{\text{orig}} \). The aim is to determine the optimal perturbation \( \Delta^* \) that maximizes the loss function \( \mathcal{L} \) when applied to \( I_{\text{orig}} \). The perturbation is restricted to a bounded domain to ensure the adversarial example \( I_{\text{adv}} \) remains within a permissible visual deviation from \( I_{\text{orig}} \). The optimal perturbation \( \Delta^* \) is found by solving:
\begin{equation}
\Delta^* = \arg\max_{\Delta} \mathcal{L}(f_{\text{white}}(I_{\text{orig}})), \label{con:goal}
\end{equation}
subject to the constraint:
\begin{equation}
||\Delta||_p \leq \epsilon,
\end{equation}
where \( ||\cdot||_p \) represents the \( p \)-norm, and \( \epsilon \) is the perturbation bound. Once \( \Delta^* \) is obtained, the adversarial example is computed as:
\begin{equation}
I_{\text{adv}} = I_{\text{orig}} + \Delta^*.
\end{equation}

Our goal with respect to a black-box model \( M_{\text{black}} \) characterized by an encoding function \( f_{\text{black}} \), is to ensure that the adversarial example \( I_{\text{adv}} \) deviates from the original image \( I_{\text{orig}} \) by at least a threshold \( \theta \) in the feature space of \( f_{\text{black}} \). This criterion for successful misclassification can be defined as follows:
\begin{equation}
|| f_{\text{black}}(I_{\text{adv}}) - f_{\text{black}}(I_{\text{orig}}) ||_2 \geq \theta,
\end{equation}
where \( ||\cdot||_2 \) denotes the Euclidean norm, signifying the measure of dissimilarity between the feature representations of the adversarial and original examples within the black-box model's domain. This condition implies that the adversarial example \( x_{\text{adv}} \) is sufficiently distinct from the original image \( x_{\text{orig}} \) to affect the black-box model's output, potentially leading to incorrect classification.

\textbf{Adversary’s capabilities.} We postulate that the adversary possesses a comprehensive understanding of the white-box model, denoted as \( M_{\text{white}} \). This encompasses an intimate familiarity with the model's architectural framework, its parameter configurations, and the gradients that dictate learning. In stark contrast to the transparency afforded by \( M_{\text{white}} \), the adversary's purview over the black-box model, \( M_{\text{black}} \), is notably opaque. Devoid of any direct knowledge regarding the internal mechanisms, parameter values, or gradient information of \( M_{\text{black}} \), the adversary operates without the capacity to compute internal representations or gradients. 

\subsection{Observation \& Analysis}

The Set-level Guidance Attack (SGA) method employs a comprehensive approach, focusing on sets of image-text pairs and utilizing all textual descriptions alongside a wide array of image augmentations for generating adversarial examples. This process follows a cyclic methodology, where adversarial text is initially guided by original images, followed by the generation of adversarial images influenced by this text, and concludes with the creation of further adversarial text influenced by these images.

Figure \ref{fig:overfitting} demonstrates the impact of varying scaling factors on the success rate of image-text matching attacks. In this setup, ALBEF is used as the source model and TCL as the target model. The observed trend, where success rates increase initially with scaling factors but subsequently decrease, suggests that while augmentation initially improves example diversity, excessive scaling leads to overfitting.

Our analysis of this phenomenon involves two key aspects. \textbf{First}, we examine overfitting during the phase where adversarial images are created based on text. In this phase, both original images and adversarial texts are encoded, resulting in unique feature representations. The success of adversarial examples relies on how closely these features match. However, using too much data augmentation can disrupt this match, leading to adversarial examples that either miss or exaggerate certain image features. This can make the attacks less effective, particularly if they focus too much on prominent features while neglecting finer details. \textbf{Second}, to address this, our approach balances data augmentation and intermodal guidance more effectively. We aim for a better match between modified examples and texts, enhancing the adaptability of adversarial examples to different models and reducing the risk of overfitting, especially in cases where subtle or less prominent image features are involved.

\subsection{The proposed Method}

\subsubsection{Symbol Conventions}

To facilitate expression, we initially agree upon certain primary notations as shown in TABLE \ref{tab:notation}.

\input{table/notation}

Given an original set of images \( \mathcal{I} \) and a set of scaling factors \( \mathcal{A} \), the augmented image set \( \mathcal{I}_{aug} \) can be obtained by first applying a horizontal flipping function \( f_{\text{flip}} \) to each image \( I \in \mathcal{I} \), followed by a scaling function \( f_{\text{scale}} \) for each scaling factor \( \alpha \in \mathcal{A} \):

\begin{equation}
\mathcal{I}_{aug} = \bigcup_{\alpha \in \mathcal{A}} \left( f_{\text{scale}}(f_{\text{flip}}(I), \alpha) \right)
\end{equation}
where \( f_{\text{flip}} \) horizontally flips the image, and \( f_{\text{scale}} \) applies the scaling transformation to the image, for each \( \alpha \).Given the augmented image set \( \mathcal{I}_{aug} \) and the original text set \( \mathcal{T} \), we apply the image encoder \( \phi \) and the text encoder \( \varphi \) to obtain the corresponding feature representations:

\begin{equation}
\mathbf{F}_{img} = \phi(\mathcal{I}_{aug})
\end{equation}

\begin{equation}
\mathbf{X}_{txt} = \varphi(\mathcal{T})
\end{equation}
where \( \mathbf{F}_{img} \) represents the features of the augmented images, and \( \mathbf{X}_{txt} \) represents the features of the original texts. The function \( \phi \) symbolizes the operation of the image encoder and \( \varphi \) the operation of the text encoder.Then the similarity matrix \( \mathbf{S} \) can be calculated as:

\begin{equation}
\mathbf{S} = \mathbf{F}_{img} \odot \mathbf{X}_{txt}
\end{equation}
where \( \mathbf{S} \) represents the similarity matrix. The operation \( \odot \) denotes matrix multiplication, which computes the similarity between each pair of image and text features.

In previous research, the direct application of the matrix \( \mathbf{S} \) to guide the generation of adversarial examples has, after several iterations, led to the reinforcement of the most feature-similar regions within the purview of the white-box model during the adversarial example creation process, with minimal perturbation occurring in other regions. When these adversarial examples encounter a new black-box model, the focal points of concern differ from those of the white-box model. Since the regions of interest to the black-box model have undergone only limited perturbation, they fail to disrupt the black-box model, resulting in an unsuccessful attack.

\begin{equation}
loss_{ori} = -\left( \sum_{i} \mathbf{S}_i \right)_{\text{mean}}
\end{equation}
Here, the summation \( \sum_{i} \mathbf{S}_i \) is taken over the last dimension of the similarity matrix \( \mathbf{S} \), and the mean of this sum is computed to obtain the final loss value \( loss_{ori} \).

\subsubsection{Optimal Transport}

\textbf{Defining Source and Target Distributions.} Initially, we define two pivotal distributions within the Optimal Transport framework: the source distribution \( \mathbf{P} \) and the target distribution \( \mathbf{Y} \). These distributions represent the starting and ending points of the transportation journey in the Optimal Transport problem. The source distribution \( \mathbf{P} = (p_1, p_2, \ldots, p_n) \) and the target distribution \( \mathbf{Y} = (y_1, y_2, \ldots, y_m) \) encapsulate the quantities to be transported from and to each respective location.

\textbf{The Transportation Matrix \( \mathbf{T} \).} In the context of Optimal Transport, the matrix \( \mathbf{T} = [T_{ij}] \) of size \( n \times m \) is referred to as the transportation matrix. Each element \( T_{ij} \) of this matrix represents the amount of a commodity or resource transported from the \( i \)-th source in \( \mathbf{P} \) to the \( j \)-th target in \( \mathbf{Y} \). The matrix \( \mathbf{T} \) effectively captures the transportation scheme between the sources and targets.

The matrix \( \mathbf{T} \) must satisfy specific constraints to ensure an optimal transportation plan. Firstly, the Marginal Constraints:
\begin{align}
\sum_{j=1}^{m} T_{ij} &= p_i \quad \forall i \in \{1, \ldots, n\}, \\
\sum_{i=1}^{n} T_{ij} &= y_j \quad \forall j \in \{1, \ldots, m\}.
\end{align}
These constraints require that the total transported amount from each source \( i \) and to each target \( j \) matches the respective supply \( p_i \) and demand \( y_j \).

Additionally, the Non-Negativity Constraint is imposed:
\begin{equation}
T_{ij} \geq 0 \quad \forall i \in \{1, \ldots, n\}, \forall j \in \{1, \ldots, m\}.
\end{equation}
This condition ensures that all transport amounts \( T_{ij} \) in the matrix \( \mathbf{T} \) are non-negative, reflecting the practical impossibility of negative transportation.

\textbf{Modeling the Optimal Transport Problem.} With the aforementioned definitions and constraints established, the Optimal Transport problem can be formulated as follows:
\begin{equation}
OT(\mathbf{P}, \mathbf{Y}, \mathbf{C}) = \min_{\mathbf{T} \in \Pi(\mathbf{r}, \mathbf{c})} \sum_{i,j} T_{ij}C_{ij}
\end{equation}
Here, \( \mathbf{C} \) denotes the cost matrix, with each element \( C_{ij} \) representing the cost of transporting a unit from source \( p_i \) to target \( y_j \). The matrix \( \mathbf{T} \) signifies the transportation scheme, while \( \Pi(\mathbf{r}, \mathbf{c}) \) encompasses all feasible transportation schemes that satisfy the marginal constraints.

The Sinkhorn distance is utilized in Optimal Transport (OT) for its effectiveness in high-dimensional spaces. Traditional OT approaches, based on linear programming, face challenges with computational intensity and scaling with data dimensionality. In contrast, the Sinkhorn distance applies entropy regularization to the OT calculation, enhancing tractability and differentiability. This approach uses a regularization parameter \( \lambda \), which balances accuracy and computational efficiency. Higher \( \lambda \) values lead to results closer to traditional OT but at increased computational costs, while lower values of \( \lambda \) expedite calculations at the expense of some bias. Therefore, the Sinkhorn distance, often computed using the Sinkhorn-Knopp algorithm, presents a more feasible solution for OT in machine-learning scenarios that demand scalability and stability in computations.
The Sinkhorn Optimization Process can be defined as:
\begin{equation}
OT_\lambda(\mathbf{P}, \mathbf{Y}, \mathbf{C}) = \min_{\mathbf{T} \in \Pi(\mathbf{r}, \mathbf{c})} \sum_{i,j} T_{ij}C_{ij} + \lambda H(\mathbf{T})
\end{equation}

The Sinkhorn algorithm iteratively normalizes the rows and columns of the transport matrix to satisfy the marginal constraints while minimizing the regularized objective function\cite{cuturi2013sinkhorn}. Here, \( H(\mathbf{T}) \) is the entropy of the transport matrix, introducing regularization (controlled by \( \lambda \)) to ensure numerical stability and efficient computation. Regarding the computation of Sinkhorn,
the algorithm of the proposed OT-Attack is summarized in Algorithm \ref{alg:sinkhorn}.

\input{algorithm/sinkhorn}

\subsubsection{Calculating Loss through Optimal Transport}

The Optimal Transport loss \( \text{loss}_{OT} \) is computed using the feature representations of augmented images \( \mathbf{F}_{img} \), original texts \( \mathbf{X}_{txt} \), and the similarity matrix \( \mathbf{S} \).

\input{algorithm/text_guidance_attack}

Firstly, the cost matrix \( \mathbf{C} \) is defined as \( \mathbf{C} = 1 - \mathbf{S} \), where \( \mathbf{S} \) is the similarity matrix. This transformation converts similarity scores into a cost structure. Subsequently, we compute the exponentiated negative cost matrix \( \mathbf{K} \) for the Sinkhorn iterations, defined as \( \mathbf{K} = \exp\left(-\frac{\mathbf{C}}{\lambda}\right) \), with \( \lambda \) being a small positive regularization parameter. The Optimal Transport loss is calculated as:
\begin{equation}
\text{loss}_{OT} = \sum_{i,j} T_{ij}C_{ij}
\end{equation}
where \( T_{ij} \) in \( \mathbf{T} \) represents the optimal 'transport' of features from the \( i \)-th element in \( \mathbf{F}_{img} \) to the \( j \)-th element in \( \mathbf{X}_{txt} \), and \( C_{ij} \) is the corresponding cost in \( \mathbf{C} \).

This formulation of \( \text{loss}_{OT} \) captures the minimal cost of aligning the feature representations of the augmented images with the original texts, facilitating a more effective generation of adversarial examples. In essence, this approach for computing OT loss in adversarial example generation accounts for the overall feature distribution, addressing potential overfitting issues inherent in using a similarity matrix as the sole loss metric. It ensures that adversarial examples generated through this method exhibit better transferability to novel black-box models. The process of generating adversarial images is detailed in Algorithm \ref{alg:text_guidance_attack}.

We employed the adversarial example generation method outlined in Equation \ref{con:goal} to create adversarial samples. These samples were then used to mount attacks on black-box models.

%% file: table/notation.tex
\begin{table}[ht]
\centering
\begin{tabular}{>{$}l<{$} l}
\hline
\text{Symbol} & \text{Description} \\
\hline
\mathcal{I} & Original Image Set \\
\mathcal{I}_{aug} & Augmented Image Set \\
\phi & Image Encoder \\
\mathbf{F}_{img} & Features of Augmented Images \\
\mathcal{T} & Text Set \\
\varphi & Text Encoder \\
\mathbf{X}_{txt} & Text Features \\
\hline
\end{tabular}
\caption{Symbol conventions.}
\label{tab:notation}
\end{table}

%% file: algorithm/sinkhorn.tex
\begin{algorithm}
\caption{Sinkhorn Iteration for Optimal Transport}
\label{alg:sinkhorn}
\begin{algorithmic}[1]
\Require $K$: cost matrix, $u$: source measure, $v$: target measure
\Ensure $T$: transport matrix

\State $r \gets \text{ones\_like}(u)$
\State $c \gets \text{ones\_like}(v)$
\State $thresh \gets 1e-2$
\For{$i = 1, \dots, 100$}
    \State $r_0 \gets r$
    \State $r \gets u / (\text{MatMul}(K, c))$
    \State $c \gets v / (\text{MatMul}(K^\top, r))$
    \State $err \gets \text{Mean}(\text{Abs}(r - r_0))$
    \If{$err < thresh$}
        \State \textbf{break}
    \EndIf
\EndFor
\State $T \gets \text{Outer}(r, c) \times K$
\State \Return $T$
\end{algorithmic}
\end{algorithm}

%% file: algorithm/text_guidance_attack.tex
\begin{algorithm}
\caption{Adversarial Image Generation}
\label{alg:text_guidance_attack}
\begin{algorithmic}[1]
\Require $model$: model for attack, $imgs$: input images, $device$: computation device, $\alpha$: scaling factors, $X_{txt}$: text embeddings 
\Ensure $I_{adv}$: adversarial images

\State $model.eval()$
\State Initialize $I_{adv} \gets imgs.detach() + \text{Uniform}(-\epsilon, \epsilon)$ and clamp to $[0.0, 1.0]$
\For{each iteration $i = 1$ to $N$}
    \For{each $img \in I_{adv}$}
        \State Apply transformations and data augmentation to $img$
        \State Select text embeddings for operation
        \State Compute similarity and Wasserstein distance
        \State Perform Sinkhorn optimization to obtain $T$
        \If{$\text{isnan}(T)$} 
            \State \Return $None$
        \EndIf
        \State Compute and backpropagate loss $loss_{OT}$
        \State Update adversarial image using gradient sign
        \State Clamp $I_{adv}'$ within perturbation limits
        \State Update $I_{adv}$ with the new adversarial image
    \EndFor
\EndFor
\State \Return $I_{adv}$
\end{algorithmic}
\end{algorithm}

%% file: section/Experiments.tex
\section{Experiments}

\input{table/result_flick30k}

\input{table/result_mscoco}

\begin{figure*}
    \centering
    \includegraphics[width=0.85\textwidth]{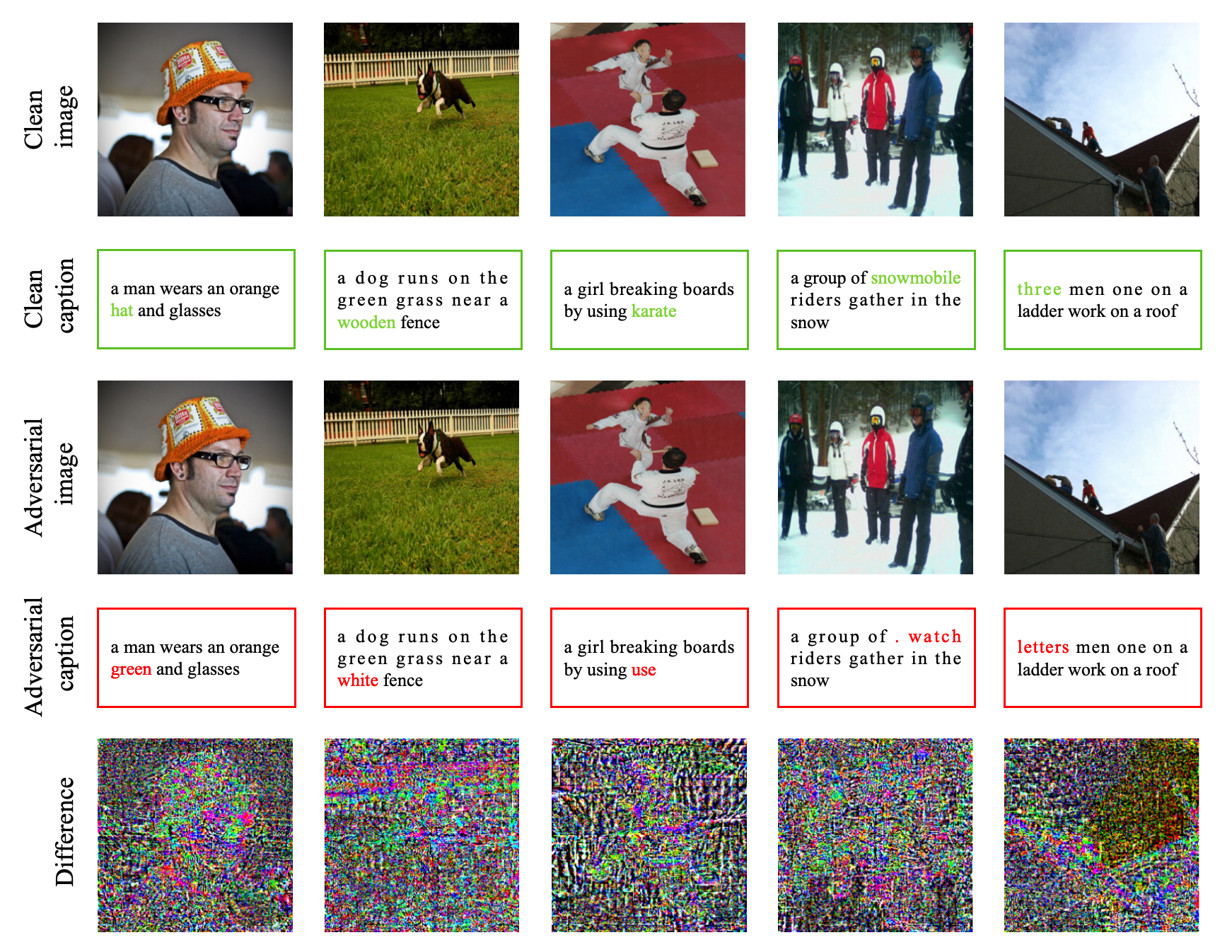}
    \caption{Visualization of adversarial examples from Flickr30K. In the task of image-text matching, adversarial examples for both images and texts were generated and utilized for image-to-text and text-to-image matching tasks, respectively. We have highlighted the distinctions in the text adversarial examples compared to the original samples and also quantified the pixel differences between the image adversarial examples and the original images.}
    \label{fig:Visualization}
\end{figure*}

\subsection{Settings}
\textbf{VLP Models.}
To assess the transferability of adversarial examples and the efficacy of our proposed framework, we employed two distinct categories of Vision-Language Pre-training (VLP) models: fused VLP and aligned VLP. These were respectively designated as the source and target models in our experiments. Fused VLP models are characterized by their early integration of visual and linguistic information within the processing pipeline. Specifically, these models concurrently handle images and text, extracting and processing both types of data through shared layers. Within this category, we selected ALBEF\cite{li2021align} and TCL\cite{yang2022vision} as representative fused models. Both models incorporate a 12-layer ViT-B/16\cite{dosovitskiy2020image} visual transformer and two separate 6-layer transformers for image and text encoding, with their primary differences arising from distinct pre-training objectives. Conversely, aligned VLP models process visual and textual data independently at the initial stages and subsequently align these representations in the deeper layers of the model. This approach facilitates the learning of intricate relationships between the two modalities. In this category, we opted for two variants of CLIP\cite{radford2021learning} as our representatives: CLIP\textsubscript{ViT}, which employs ViT-B/16 for image encoding, and CLIP\textsubscript{CNN}, which uses a ResNet-101\cite{he2016deep} architecture for the same purpose. For the targeted task of image captioning within our cross-task attack efficacy study, we incorporated BLIP as the target model, assessing adversarial examples' performance in the black-box setting when transitioning from TCL as the source model to ALBEF.

\textbf{Datasets.}
For the image-text retrieval task, our study utilized two datasets renowned for their breadth and depth: Flickr30K\cite{plummer2015flickr30k} and MSCOCO\cite{lin2014microsoft}. Flickr30K boasts a diverse corpus of 31,783 images, while MSCOCO expands the dataset considerably with 123,287 images. A salient characteristic shared by both is the quintuple of descriptive captions accompanying each image, providing a valuable asset for the assessment of our image-text retrieval approach. For the task of Visual Grounding, we employed the RefCOCO+\cite{yu2016modeling} dataset, which further enriched our cross-task attack effectiveness analysis.

\textbf{Baselines}
In our research involving Vision-Language Pre-training (VLP) models, we implemented several prevalent adversarial attack methods as baselines. These included using PGD\cite{madry2017towards} exclusively on images, applying BERT-Attack\cite{li2020bert} only to texts, and separately utilizing PGD and BERT-Attack on both images and texts without integrating inter-modality interactions, a technique designated as Sep-Attack. Additionally, we employed Co-Attack\cite{zhang2022towards}, which integrates information between individual image-text pairs, and Set-level Guidance Attack (SGA)\cite{lu2023set}, which utilizes guidance information across modalities between sets. Each baseline was tested under identical conditions for a consistent comparative analysis.

\textbf{Adversarial Attack Configuration.} 
In order to validate the effectiveness of our proposed framework, we employed the Projected Gradient Descent (PGD)\cite{madry2017towards} method for generating adversarial visual examples. The configuration settings for PGD were as follows: a perturbation limit of $\epsilon_v = \frac{2}{255}$, a step magnitude of $\alpha = \frac{0.5}{255}$, and a total of $T = 10$ iterations. For textual attacks, we utilized the BERT-Attack\cite{li2020bert} method, setting a disturbance limit to $\epsilon_t = 1$ and defining a vocabulary list length of $W = 10$. In our experimentation with Sep-Attack and Co-Attack, we adhered to the previously mentioned settings. Specifically for Co-Attack, in addition to these settings, we also incorporated the similarity between individual image pairs as a loss metric, guiding the generation of adversarial examples through inter-modality interactions. In the case of SGA, we followed the experimental conditions outlined in its original publication. Specifically, we enhanced the images by rescaling them to five distinct sizes ${0.5, 0.75, 1.0, 1.25, 1.5}$. To validate that our method effectively mitigates overfitting, we enhanced the image set by rescaling it to six different sizes: ${0.5, 0.75, 1.0, 1.25, 1.5, 2}$, utilizing bicubic interpolation. Additionally, we applied horizontal flipping to the images. This form of data augmentation, which led to significant overfitting in SGA, was employed to rigorously test the robustness of our approach against such overfitting tendencies. In parallel, the text corpus was expanded by selecting and refining the top five most closely matched caption pairs for each image in the dataset. Furthermore, we incorporated the Sinkhorn\cite{cuturi2013sinkhorn} algorithm for calculating the optimal transport plan. This algorithm was chosen for its efficient use of matrix operations, which accelerates the process and ensures convergence to an approximate solution of the original transport problem. To avoid infinite iterations, we set a convergence threshold $\text{thresh} = 1e-2$. The iteration process was terminated once the average absolute difference of the vector $r$ between two consecutive iterations fell below this threshold, indicating the achievement of algorithmic convergence.

\textbf{Evaluation Criteria.} 
In our study, the robustness and transferability of the adversarial attacks, in both white-box and black-box scenarios, are quantitatively assessed using the Attack Success Rate (ASR). ASR is a crucial metric that measures the proportion of successful adversarial examples out of the total number of attacks conducted. A higher ASR is indicative of increased transferability of the adversarial examples, signifying the effectiveness of the attack in compromising the model under various conditions. The ASR is computed as follows:

\begin{equation}
    ASR = \frac{N_{\text{success}}}{N_{\text{total}}} \times 100\%
\end{equation}
where \( ASR \) denotes the Attack Success Rate, \( N_{\text{success}} \) represents the number of successful attacks, and \( N_{\text{total}} \) is the total number of attacks conducted. The formula calculates the percentage of successful attacks, providing a quantitative measure of the attack's effectiveness.

\subsection{Comparative Experimental Results}

In our experiments, we primarily focused on Image-Text Retrieval (ITR) tasks. We generated adversarial examples on various white-box models and then evaluated their effectiveness by calculating the attack success rates on both the white-box models and three additional black-box models. Specifically, we computed the success rates for both image-to-text and text-to-image matching attacks to demonstrate the comprehensiveness of our method's impact on adversarial sample generation. This deliberate selection of models and tasks was instrumental in enabling an exhaustive examination of the adversarial examples' ability to generalize across varying architectures, which is a critical factor in the real-world application of these models.

Our analysis spanned two widely recognized datasets: Flickr30K, with a sample of 1,000 images and 5,000 captions, and MSCOCO, which provided a larger pool of 5,000 images and 25,000 captions. This broad dataset coverage allowed us to conduct a robust evaluation of our attack methods in image-text matching tasks, quantifying the success of adversarial examples in misleading these complex models. The detailed outcomes are methodically presented in TABLE \ref{tab:result_flicker30K} and TABLE \ref{tab:result_mscoco}.

Our results demonstrated that the OT-Attack method made significant strides in the creation of adversarial examples that were not only effective within models of the same type but also exhibited impressive cross-type attack success. This is particularly evident from the R@1 success rates in TR and IR tasks, where our adversarial examples maintained high effectiveness across varied models including ALBEF, TCL, CLIP\textsubscript{ViT}, and CLIP\textsubscript{CNN}. For example, when using ALBEF to target TCL, our method improved the TR R@1 attack success rate by 6.95\% on Flickr30K and 4.88\% on MSCOCO, compared with the state-of-the-art results obtained by SGA. Conversely, in scenarios where TCL was employed to target ALBEF, our approach showed significant improvements over SGA, with increases of 8.41\% on Flickr30K and 5.71\% on MSCOCO in the TR R@1 attack success rate. The results demonstrate the effectiveness of our proposed method in improving adversarial transferability.



Complementing our numerical analysis, Figure \ref{fig:Visualization} offers a visual representation of the impact of our adversarial examples. It contrasts the original images and texts with their modified versions, illuminating how subtle perturbations can drastically alter a model's performance in image-text matching tasks. The visual differences, particularly the nuanced texture changes introduced in the adversarial images, are made evident through difference masks, underscoring the deceptive potency of the adversarial examples and their potential to misguide VLP systems. This visual depiction reinforces the quantitative data, providing a holistic view of our adversarial attack's efficacy. Hence, compared with previous works, our proposed OT-Attack can achieve better adversarial transferability.

\subsection{Cross-Task Transferability}

\subsubsection{Image Captioning}

\begin{figure*}
    \centering
    \includegraphics[width=0.8\textwidth]{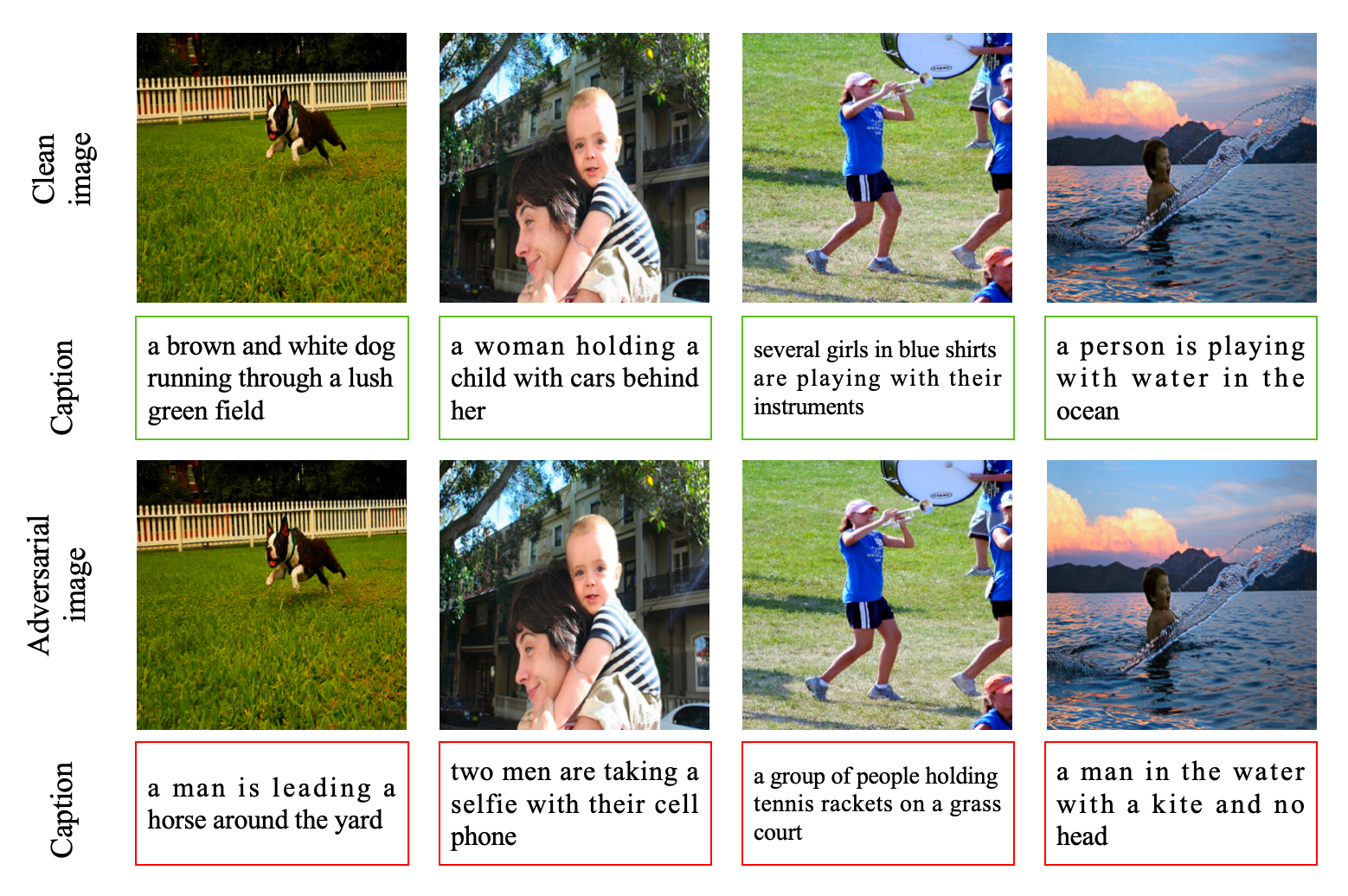}
    \caption{Comparison of Clean and Adversarial Image Captions. This figure juxtaposes the original clean images with their accurate captions against adversarial images and the resulting captions generated by the BLIP model. The adversarial examples were created using the ALBEF model as a white-box framework on the Dataset Flickr30K. Despite the perturbations being subtle, and limited to a magnitude of 2, the adversarial examples show minimal visual deviation from the original images. However, these slight alterations are significant enough to mislead the captioning model, leading to discrepancies in the generated captions, as evidenced by the erroneous and sometimes nonsensical descriptions.}
    \label{fig:caption}
\end{figure*}

\input{table/caption}

\begin{figure*}
    \centering
    \includegraphics[width=0.95\textwidth]{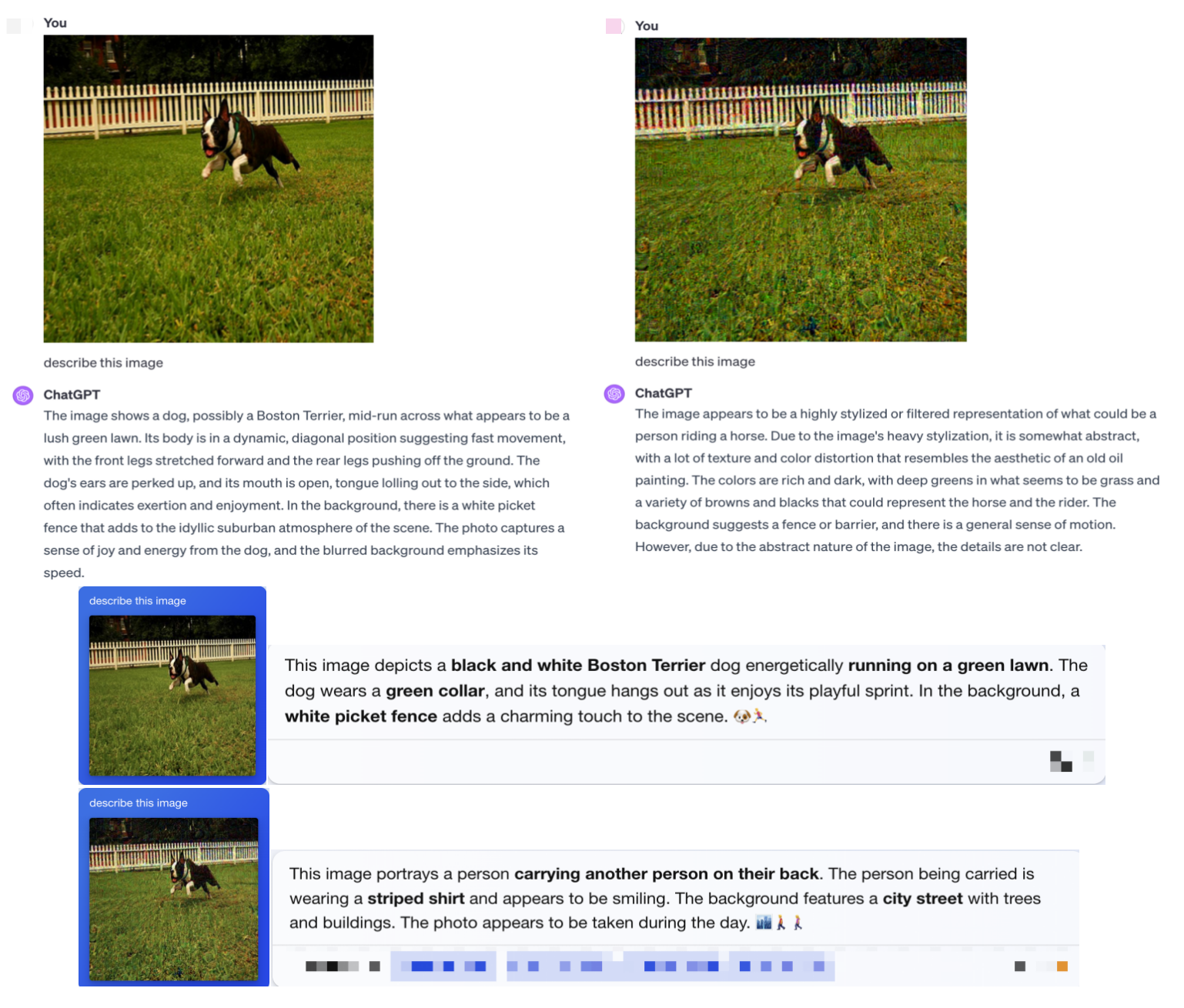}
    \caption{Impact of Adversarial Attacks on GPT-4 and Bing Chat Descriptions. This figure showcases the alterations in image descriptions by GPT-4 and Bing Chat before and after adversarial attacks. Original descriptions are compared to those generated from manipulated images, with increased perturbation strength and iteration count to mislead the AI models. The stark contrast in the outputs highlights the susceptibility of these models to adversarial examples, reflecting the effectiveness of the perturbations in altering the perceived content of the images.}
    \label{fig:gpt}
\end{figure*}

In our research, we employed the ALBEF model in a white-box setting to generate adversarial examples specifically designed to target the BLIP\cite{li2022blip} framework. Renowned for its innovative multimodal encoder-decoder model, BLIP is pre-trained on a dataset enriched with diverse synthetic captions and noise reduction techniques. We chose the MSCOCO dataset for our evaluations, focusing on both original and adversarially manipulated image samples. To quantify the effectiveness of our adversarial approach, we utilized a suite of metrics for image captioning tasks, each offering distinct insights: \textbf{BLEU\cite{papineni2002bleu}} assesses the precision of generated captions by comparing them with reference captions, focusing on the exactness of word usage and sequence. \textbf{METEOR\cite{banerjee2005automatic}} expands upon BLEU by including synonym matching and stemming, providing a more comprehensive view of semantic accuracy. \textbf{ROUGE\cite{lin2004rouge}} measures recall, evaluating how much of the reference content is captured in the generated captions. \textbf{CIDEr\cite{vedantam2015cider}} focuses on the distinctiveness and relevance of captions by assessing consensus with reference captions. Lastly, \textbf{SPICE\cite{anderson2016spice}} evaluates the semantic content of captions, assessing their fidelity to the actual objects, attributes, and relationships in the image.

Each metric provides a different perspective on the quality and relevance of the generated text, offering a comprehensive understanding of the adversarial impact. The detailed results of these evaluations are compiled in TABLE \ref{tab:caption}. Compared to SGA, our method exhibited a decrease in performance across various metrics: a reduction of 0.7 in BLEU-4, 0.5 in METEOR, 0.6 in ROUGE-L, 3.4 in CIDEr, and 0.5 in SPICE. Notably, these declines in scores are indicative of enhanced effectiveness in cross-task attacks, as a greater decrease signifies better performance of our method in this context.

To convey the practical implications of our findings, Figure \ref{fig:caption} provides a visual comparison of select experimental results. This figure juxtaposes the original images and their corresponding captions with the adversarial counterparts, highlighting the stark differences. These visual examples compellingly illustrate how subtle and seemingly innocuous perturbations can lead to significant deviations in the model's interpretation, often diverging entirely from the original context or meaning of the image. 

Further delving into the realm of large-scale models, our experiments were conducted with specific parameters to gauge the extent of adversarial impact. We set the perturbation intensity at a subtle yet effective level of 16/255 and ran our adversarial process for 500 iterations. To assess the broader applicability and effectiveness of our attacks, we tested them on advanced models like GPT-4 and Bing Chat, posing the query ``Describe this image" to these systems. The findings, illustrated in Figure \ref{fig:gpt}, reveal a notable level of success in our adversarial attacks, with these sophisticated models showing susceptibility to being misled. 

\input{table/visual_grounding}

\subsubsection{Visual Grounding}

\begin{figure*}
    \centering
    \includegraphics[width=0.8\textwidth]{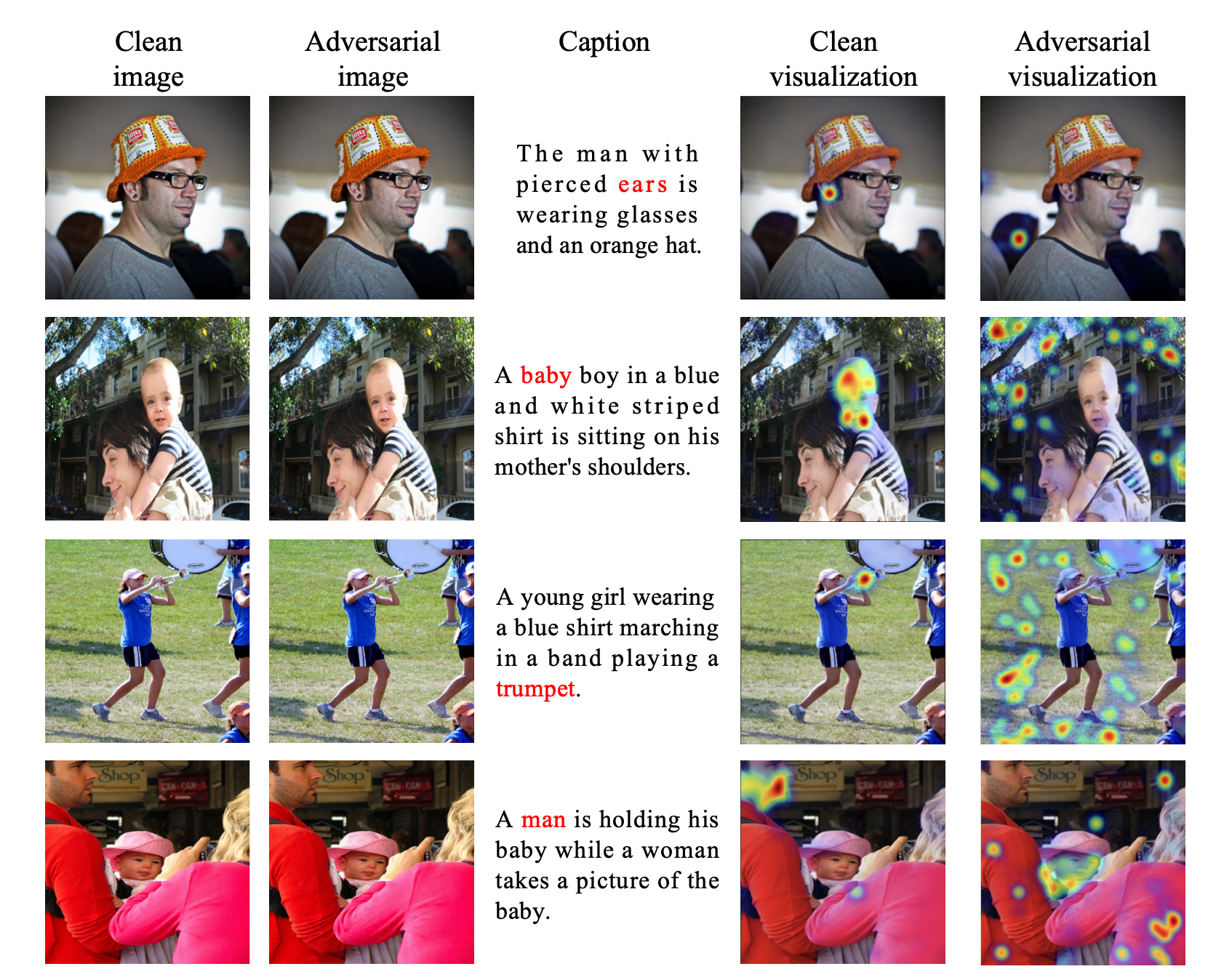}
    \caption{Visualization results for Visual Grounding. We employed TCL as the source model and ALBEF as the target model, with captions sourced from the Flickr30K dataset. The adversarial examples exhibit limited visual differences from the original samples; however, they disrupt the model's judgment of visual elements in the Visual Grounding task. Compared to clean data, the localization results for the same elements may have shifted or dispersed. The visualizations of Visual Grounding vividly demonstrate the disruptive impact of adversarial examples on the model.}
    \label{fig:vg}
\end{figure*}

To thoroughly evaluate the efficacy of our adversarial attack strategies, we utilized the RefCOCO+\cite{yu2016modeling} dataset, meticulously designed for the visual grounding task. This dataset is composed of several subsets, each tailored to assess different facets of model performance. These subsets include RefCOCO+ val, RefCOCO+ testA, and RefCOCO+ testB. Specifically, the RefCOCO+ val subset presents a wide array of scenarios for a comprehensive assessment, RefCOCO+ testA is dedicated to gauging the model's proficiency in identifying and localizing human figures, and RefCOCO+ testB concentrates on non-human elements like inanimate objects, animals, and various environmental features. By leveraging the diverse testing environments provided by RefCOCO+, our goal is to showcase the extensive adaptability and transferability of our method across a multitude of visual grounding challenges.

The quantitative analysis presented in TABLE \ref{tab:visual_grounding} assesses the performance of our adversarial examples across different tasks. In this evaluation, TCL served as the source model from which adversarial examples were generated, while the ALBEF model was the target for these attacks. The baseline in the table refers to the scores obtained when the original, unaltered samples were evaluated, providing a control benchmark for our study. Alongside the baseline, both SGA and OT-Attack strategies were employed. Compared to SGA, our attack method proved to be more effective, as evidenced by the decrease in scores for the ALBEF model due to our adversarial examples: a reduction of 0.2 on the Val subset, 0.2 on TestA, and 0.3 on TestB. This outcome highlights the effectiveness of our method in compromising the Visual Grounding capabilities of the target model.

Beyond the numerical assessments, we also employed a visualization technique to further elucidate the impact of our attacks on Visual Grounding tasks, as depicted in Figure \ref{fig:vg}. The base samples for this analysis were derived from the Flickr30K dataset, with adversarial examples being crafted using the ALBEF model in a white-box configuration. Our visual representations reveal that even subtle perturbations can significantly disrupt the model's ability to accurately recognize and localize objects. This underscores the substantial and nuanced impact that these adversarial attacks can have on a model's performance in visual grounding tasks, challenging its reliability and precision in real-world applications.

%% file: table/result_flick30k.tex
\begin{table*}[]
\centering
\begin{adjustbox}{width=\textwidth,center}
\begin{tabular}{@{}l|l|cc|cc|cc|cc@{}}
\toprule
 &
   &
  \multicolumn{2}{c|}{\textbf{ALBEF}} &
  \multicolumn{2}{c|}{\textbf{TCL}} &
  \multicolumn{2}{c|}{\textbf{CLIP\textsubscript{ViT}}} &
  \multicolumn{2}{c}{\textbf{CLIP\textsubscript{CNN}}} \\ \cmidrule(l){3-10} 
\multirow{-2}{*}{\textbf{Source}} &
  \multirow{-2}{*}{\textbf{Attack}} &
  TR R@1 &
  IR R@1 &
  TR R@1 &
  IR R@1 &
  TR R@1 &
  IR R@1 &
  TR R@1 &
  IR R@1 \\ \midrule
 &
  PGD &
  \cellcolor[HTML]{EFEFEF}52.45 &
  \cellcolor[HTML]{EFEFEF}58.65 &
  3.06 &
  6.79 &
  8.69 &
  13.21 &
  10.34 &
  14.65 \\
 &
  BERT-Attack &
  \cellcolor[HTML]{EFEFEF}11.57 &
  \cellcolor[HTML]{EFEFEF}27.46 &
  12.64 &
  28.07 &
  29.33 &
  43.17 &
  32.69 &
  46.11 \\
 &
  Sep-Attack &
  \cellcolor[HTML]{EFEFEF}65.69 &
  \cellcolor[HTML]{EFEFEF}73.95 &
  17.60 &
  32.95 &
  31.17 &
  45.23 &
  32.82 &
  45.49 \\
 &
  Co-Attack &
  \cellcolor[HTML]{EFEFEF}77.16 &
  \cellcolor[HTML]{EFEFEF}83.86 &
  15.21 &
  29.49 &
  23.60 &
  36.48 &
  25.12 &
  38.89 \\
 &
  SGA &
  \cellcolor[HTML]{EFEFEF}97.24 &
  \cellcolor[HTML]{EFEFEF}97.28 &
  45.42 &
  55.25 &
  33.38 &
  44.16 &
  34.93 &
  46.57 \\
\multirow{-6}{*}{\textbf{ALBEF}} &
  \cellcolor[HTML]{C0C0C0}OT-Attack (Ours) &
  \cellcolor[HTML]{C0C0C0}95.93 &
  \cellcolor[HTML]{C0C0C0}95.86 &
  \cellcolor[HTML]{C0C0C0}\textbf{52.37} &
  \cellcolor[HTML]{C0C0C0}\textbf{61.05} &
  \cellcolor[HTML]{C0C0C0}\textbf{34.85} &
  \cellcolor[HTML]{C0C0C0}\textbf{47.10} &
  \cellcolor[HTML]{C0C0C0}\textbf{42.33} &
  \cellcolor[HTML]{C0C0C0}\textbf{53.03} \\ \midrule
 &
  PGD &
  6.15 &
  10.78 &
  \cellcolor[HTML]{EFEFEF}77.87 &
  \cellcolor[HTML]{EFEFEF}79.48 &
  7.48 &
  13.72 &
  10.34 &
  15.33 \\
 &
  BERT-Attack &
  11.89 &
  26.82 &
  \cellcolor[HTML]{EFEFEF}14.54 &
  \cellcolor[HTML]{EFEFEF}29.17 &
  29.69 &
  44.49 &
  33.46 &
  46.07 \\
 &
  Sep-Attack &
  20.13 &
  36.48 &
  \cellcolor[HTML]{EFEFEF}84.72 &
  \cellcolor[HTML]{EFEFEF}86.07 &
  31.29 &
  44.65 &
  33.33 &
  45.80 \\
 &
  Co-Attack &
  23.15 &
  40.04 &
  \cellcolor[HTML]{EFEFEF}77.94 &
  \cellcolor[HTML]{EFEFEF}85.59 &
  27.85 &
  41.19 &
  30.74 &
  44.11 \\
 &
  SGA &
  48.91 &
  60.34 &
  \cellcolor[HTML]{EFEFEF}98.37 &
  \cellcolor[HTML]{EFEFEF}98.81 &
  33.87 &
  44.88 &
  37.74 &
  48.30 \\
\multirow{-6}{*}{\textbf{TCL}} &
  \cellcolor[HTML]{C0C0C0}OT-Attack (Ours) &
  \cellcolor[HTML]{C0C0C0}\textbf{57.32} &
  \cellcolor[HTML]{C0C0C0}\textbf{65.83} &
  \cellcolor[HTML]{C0C0C0}97.81 &
  \cellcolor[HTML]{C0C0C0}98.01 &
  \cellcolor[HTML]{C0C0C0}\textbf{34.72} &
  \cellcolor[HTML]{C0C0C0}\textbf{47.16} &
  \cellcolor[HTML]{C0C0C0}\textbf{43.44} &
  \cellcolor[HTML]{C0C0C0}\textbf{54.12} \\ \midrule
 &
  PGD &
  2.50 &
  4.93 &
  4.85 &
  8.17 &
  \cellcolor[HTML]{EFEFEF}70.92 &
  \cellcolor[HTML]{EFEFEF}78.61 &
  5.36 &
  8.44 \\
 &
  BERT-Attack &
  9.59 &
  22.64 &
  11.80 &
  25.07 &
  \cellcolor[HTML]{EFEFEF}28.34 &
  \cellcolor[HTML]{EFEFEF}39.08 &
  30.40 &
  37.43 \\
 &
  Sep-Attack &
  9.59 &
  23.25 &
  11.38 &
  25.60 &
  \cellcolor[HTML]{EFEFEF}79.75 &
  \cellcolor[HTML]{EFEFEF}86.79 &
  30.78 &
  39.76 \\
 &
  Co-Attack &
  10.57 &
  24.33 &
  11.94 &
  26.69 &
  \cellcolor[HTML]{EFEFEF}93.25 &
  \cellcolor[HTML]{EFEFEF}95.86 &
  32.52 &
  41.82 \\
 &
  SGA &
  13.40 &
  27.22 &
  16.23 &
  30.76 &
  \cellcolor[HTML]{EFEFEF}99.08 &
  \cellcolor[HTML]{EFEFEF}98.94 &
  38.76 &
  47.79 \\
\multirow{-6}{*}{\textbf{CLIP\textsubscript{ViT}}} &
  \cellcolor[HTML]{C0C0C0}OT-Attack (Ours) &
  \cellcolor[HTML]{C0C0C0}\textbf{14.29} &
  \cellcolor[HTML]{C0C0C0}\textbf{29.28} &
  \cellcolor[HTML]{C0C0C0}\textbf{16.58} &
  \cellcolor[HTML]{C0C0C0}\textbf{33.49} &
  \cellcolor[HTML]{C0C0C0}98.65 &
  \cellcolor[HTML]{C0C0C0}98.52 &
  \cellcolor[HTML]{C0C0C0}\textbf{43.55} &
  \cellcolor[HTML]{C0C0C0}\textbf{50.50} \\ \midrule
 &
  PGD &
  2.09 &
  4.82 &
  4.00 &
  7.81 &
  1.10 &
  6.60 &
  \cellcolor[HTML]{EFEFEF}86.46 &
  \cellcolor[HTML]{EFEFEF}92.25 \\
 &
  BERT-Attack &
  8.86 &
  23.27 &
  12.33 &
  25.48 &
  27.12 &
  37.44 &
  \cellcolor[HTML]{EFEFEF}30.40 &
  \cellcolor[HTML]{EFEFEF}40.10 \\
 &
  Sep-Attack &
  8.55 &
  23.41 &
  12.64 &
  26.12 &
  28.34 &
  39.43 &
  \cellcolor[HTML]{EFEFEF}91.44 &
  \cellcolor[HTML]{EFEFEF}95.44 \\
 &
  Co-Attack &
  8.79 &
  23.74 &
  13.10 &
  26.02 &
  28.79 &
  40.03 &
  \cellcolor[HTML]{EFEFEF}94.76 &
  \cellcolor[HTML]{EFEFEF}96.89 \\
 &
  SGA &
  11.42 &
  24.80 &
  \textbf{14.91} &
  28.82 &
  31.24 &
  42.12 &
  \cellcolor[HTML]{EFEFEF}99.24 &
  \cellcolor[HTML]{EFEFEF}99.49 \\
\multirow{-6}{*}{\textbf{CLIP\textsubscript{CNN}}} &
  \cellcolor[HTML]{C0C0C0}OT-Attack (Ours) &
  \cellcolor[HTML]{C0C0C0}\textbf{11.57} &
  \cellcolor[HTML]{C0C0C0}\textbf{26.24} &
  \cellcolor[HTML]{C0C0C0}\textbf{14.91} &
  \cellcolor[HTML]{C0C0C0}\textbf{30.52} &
  \cellcolor[HTML]{C0C0C0}\textbf{35.63} &
  \cellcolor[HTML]{C0C0C0}\textbf{48.20} &
  \cellcolor[HTML]{C0C0C0}99.39 &
  \cellcolor[HTML]{C0C0C0}99.32 \\ \bottomrule
\end{tabular}
\end{adjustbox}
\caption{Adversarial Attack Success Rates on Image-Text Retrieval. \textbf{The best results are boldfaced}. The table presents the attack success rate at Rank 1 (ASR @ R1) for text-image retrieval (IR) and text-image retrieval (TR) tasks using the Flickr30K dataset. The originating models for generating adversarial samples are listed under the `Source' column. White-box attacks are distinguished by a gray background, emphasizing their distinct context in the evaluation. \textbf{The results demonstrate the proposed OT-Attack's impressive performance in achieving SOTA transferability of adversarial samples across different VLP models.}}
\label{tab:result_flicker30K}
\end{table*}

%% file: table/result_mscoco.tex
\begin{table*}[]
\centering
\begin{adjustbox}{width=\textwidth,center}
\begin{tabular}{@{}l|l|cc|cc|cc|cc@{}}
\toprule
 &
   &
  \multicolumn{2}{c|}{\textbf{ALBEF}} &
  \multicolumn{2}{c|}{\textbf{TCL}} &
  \multicolumn{2}{c|}{\textbf{CLIP\textsubscript{ViT}}} &
  \multicolumn{2}{c}{\textbf{CLIP\textsubscript{CNN}}} \\ \cmidrule(l){3-10} 
\multirow{-2}{*}{\textbf{Source}} &
  \multirow{-2}{*}{\textbf{Attack}} &
  TR R@1 &
  IR R@1 &
  TR R@1 &
  IR R@1 &
  TR R@1 &
  IR R@1 &
  TR R@1 &
  IR R@1 \\ \midrule
 &
  PGD &
  \cellcolor[HTML]{EFEFEF}76.70 &
  \cellcolor[HTML]{EFEFEF}86.30 &
  12.46 &
  17.77 &
  13.96 &
  23.1 &
  17.45 &
  23.54 \\
 &
  BERT-Attack &
  \cellcolor[HTML]{EFEFEF}24.39 &
  \cellcolor[HTML]{EFEFEF}36.13 &
  24.34 &
  33.39 &
  44.94 &
  52.28 &
  47.73 &
  54.75 \\
 &
  Sep-Attack &
  \cellcolor[HTML]{EFEFEF}82.6 &
  \cellcolor[HTML]{EFEFEF}89.88 &
  32.83 &
  42.92 &
  44.03 &
  54.46 &
  46.96 &
  55.88 \\
 &
  Co-Attack &
  \cellcolor[HTML]{EFEFEF}79.87 &
  \cellcolor[HTML]{EFEFEF}87.83 &
  32.62 &
  43.09 &
  44.89 &
  54.75 &
  47.3 &
  55.64 \\
 &
  SGA &
  \cellcolor[HTML]{EFEFEF}96.75 &
  \cellcolor[HTML]{EFEFEF}96.95 &
  58.56 &
  65.38 &
  57.06 &
  62.25 &
  58.95 &
  66.52 \\
\multirow{-6}{*}{\textbf{ALBEF}} &
  \cellcolor[HTML]{C0C0C0}OT-Attack (Ours) &
  \cellcolor[HTML]{C0C0C0}95.41 &
  \cellcolor[HTML]{C0C0C0}95.8 &
  \cellcolor[HTML]{C0C0C0}\textbf{63.44} &
  \cellcolor[HTML]{C0C0C0}\textbf{68.9} &
  \cellcolor[HTML]{C0C0C0}\textbf{58.79} &
  \cellcolor[HTML]{C0C0C0}\textbf{65.87} &
  \cellcolor[HTML]{C0C0C0}\textbf{63.56} &
  \cellcolor[HTML]{C0C0C0}\textbf{72.16} \\ \midrule
 &
  PGD &
  10.83 &
  16.52 &
  \cellcolor[HTML]{EFEFEF}59.58 &
  \cellcolor[HTML]{EFEFEF}69.53 &
  14.23 &
  22.28 &
  17.25 &
  23.12 \\
 &
  BERT-Attack &
  35.32 &
  45.92 &
  \cellcolor[HTML]{EFEFEF}38.54 &
  \cellcolor[HTML]{EFEFEF}48.48 &
  51.09 &
  58.8 &
  52.23 &
  61.26 \\
 &
  Sep-Attack &
  41.71 &
  52.97 &
  \cellcolor[HTML]{EFEFEF}70.32 &
  \cellcolor[HTML]{EFEFEF}78.97 &
  50.74 &
  60.13 &
  51.9 &
  61.26 \\
 &
  Co-Attack &
  46.08 &
  57.09 &
  \cellcolor[HTML]{EFEFEF}85.38 &
  \cellcolor[HTML]{EFEFEF}91.39 &
  51.62 &
  60.46 &
  52.13 &
  62.49 \\
 &
  SGA &
  65.93 &
  73.3 &
  \cellcolor[HTML]{EFEFEF}98.97 &
  \cellcolor[HTML]{EFEFEF}99.15 &
  56.34 &
  63.99 &
  59.44 &
  65.7 \\
\multirow{-6}{*}{\textbf{TCL}} &
  \cellcolor[HTML]{C0C0C0}OT-Attack (Ours) &
  \cellcolor[HTML]{C0C0C0}\textbf{71.64} &
  \cellcolor[HTML]{C0C0C0}\textbf{78.38} &
  \cellcolor[HTML]{C0C0C0}98.69 &
  \cellcolor[HTML]{C0C0C0}98.78 &
  \cellcolor[HTML]{C0C0C0}\textbf{58.64} &
  \cellcolor[HTML]{C0C0C0}\textbf{65.75} &
  \cellcolor[HTML]{C0C0C0}\textbf{63.45} &
  \cellcolor[HTML]{C0C0C0}\textbf{72.01} \\ \midrule
 &
  PGD &
  7.24 &
  10.75 &
  10.19 &
  13.74 &
  \cellcolor[HTML]{EFEFEF}54.79 &
  \cellcolor[HTML]{EFEFEF}66.85 &
  7.32 &
  11.34 \\
 &
  BERT-Attack &
  20.34 &
  29.74 &
  21.08 &
  29.61 &
  \cellcolor[HTML]{EFEFEF}45.06 &
  \cellcolor[HTML]{EFEFEF}51.68 &
  44.54 &
  53.72 \\
 &
  Sep-Attack &
  23.41 &
  34.61 &
  25.77 &
  36.84 &
  \cellcolor[HTML]{EFEFEF}68.52 &
  \cellcolor[HTML]{EFEFEF}77.94 &
  43.11 &
  49.76 \\
 &
  Co-Attack &
  30.28 &
  42.67 &
  32.84 &
  44.69 &
  \cellcolor[HTML]{EFEFEF}97.98 &
  \cellcolor[HTML]{EFEFEF}98.8 &
  55.08 &
  62.51 \\
 &
  SGA &
  33.41 &
  44.64 &
  37.54 &
  47.76 &
  \cellcolor[HTML]{EFEFEF}99.79 &
  \cellcolor[HTML]{EFEFEF}99.79 &
  58.93 &
  65.83 \\
\multirow{-6}{*}{\textbf{CLIP\textsubscript{ViT}}} &
  \cellcolor[HTML]{C0C0C0}OT-Attack (Ours) &
  \cellcolor[HTML]{C0C0C0}\textbf{35.11} &
  \cellcolor[HTML]{C0C0C0}\textbf{46.48} &
  \cellcolor[HTML]{C0C0C0}\textbf{38.52} &
  \cellcolor[HTML]{C0C0C0}\textbf{50.32} &
  \cellcolor[HTML]{C0C0C0}99.69 &
  \cellcolor[HTML]{C0C0C0}99.75 &
  \cellcolor[HTML]{C0C0C0}\textbf{62.16} &
  \cellcolor[HTML]{C0C0C0}\textbf{68.96} \\ \midrule
 &
  PGD &
  7.01 &
  10.62 &
  10.08 &
  13.65 &
  4.88 &
  10.7 &
  \cellcolor[HTML]{EFEFEF}76.99 &
  \cellcolor[HTML]{EFEFEF}84.2 \\
 &
  BERT-Attack &
  23.38 &
  34.64 &
  24.58 &
  29.61 &
  51.28 &
  57.49 &
  \cellcolor[HTML]{EFEFEF}54.43 &
  \cellcolor[HTML]{EFEFEF}62.17 \\
 &
  Sep-Attack &
  26.53 &
  39.29 &
  30.26 &
  41.51 &
  50.44 &
  57.11 &
  \cellcolor[HTML]{EFEFEF}88.72 &
  \cellcolor[HTML]{EFEFEF}92.49 \\
 &
  Co-Attack &
  29.83 &
  41.97 &
  32.97 &
  43.72 &
  53.1 &
  58.9 &
  \cellcolor[HTML]{EFEFEF}96.72 &
  \cellcolor[HTML]{EFEFEF}98.56 \\
 &
  SGA &
  31.61 &
  43 &
  34.81 &
  45.95 &
  56.62 &
  60.77 &
  \cellcolor[HTML]{EFEFEF}99.61 &
  \cellcolor[HTML]{EFEFEF}99.8 \\
\multirow{-6}{*}{\textbf{CLIP\textsubscript{CNN}}} &
  \cellcolor[HTML]{C0C0C0}OT-Attack (Ours) &
  \cellcolor[HTML]{C0C0C0}\textbf{32.9} &
  \cellcolor[HTML]{C0C0C0}\textbf{44.03} &
  \cellcolor[HTML]{C0C0C0}\textbf{36.07} &
  \cellcolor[HTML]{C0C0C0}\textbf{48.17} &
  \cellcolor[HTML]{C0C0C0}\textbf{61.14} &
  \cellcolor[HTML]{C0C0C0}\textbf{67.79} &
  \cellcolor[HTML]{C0C0C0}99.16 &
  \cellcolor[HTML]{C0C0C0}99.59 \\ \bottomrule
\end{tabular}
\end{adjustbox}
\caption{Visualization of Adversarial Examples in Image-Text Matching on the MSCOCO Dataset. \textbf{The best results are boldfaced}. It illustrates the effects of adversarial attacks on both images and their associated captions. The top row displays the original, clean images with their corresponding accurate captions. The middle row presents the adversarial images and the modified captions that resulted from the attacks. Key alterations in the text are indicated in red, showcasing the change in the description of the visual content due to the adversarial manipulation. The bottom row quantifies the pixel differences between the original and adversarial images, highlighting the subtlety of the visual perturbations.}
\label{tab:result_mscoco}
\end{table*}

%% file: table/caption.tex
\begin{table}[]
\centering
\scalebox{0.96}{
\begin{tabular}{@{}l|ccccc@{}}
\toprule
\centering
\textbf{Attack} & \multicolumn{1}{l}{B@4} & \multicolumn{1}{l}{METEOR} & \multicolumn{1}{l}{ROUGE-L} & \multicolumn{1}{l}{CIDEr} & \multicolumn{1}{l}{SPICE} \\ \midrule
Baseline  & 39.7          & 31.0          & 60.0          & 133.3          & 23.8          \\
Co-Attack & 37.4          & 29.8          & 58.4          & 125.5          & 22.8          \\
SGA       & 34.8          & 28.4          & 56.3          & 116.0          & 21.4          \\
OT-Attack (Ours)    & \textbf{34.1} & \textbf{27.9} & \textbf{55.7} & \textbf{112.6} & \textbf{20.9} \\ \bottomrule
\end{tabular}
}
\caption{Adversarial Impact on Image Captioning Metrics. This table displays the results of adversarial attacks on the image captioning task, where 10,000 MSCOCO dataset images were used. Adversarial samples were crafted using the ALBEF model in a white-box attack scenario, and captions were subsequently generated with the BLIP model. The performance of these attacks was assessed by measuring the BLEU-4 (B@4), METEOR, ROUGE-L, CIDEr, and SPICE metrics. Lower scores on these metrics indicate a more effective attack, revealing a significant deviation of the generated captions from the expected, accurate descriptions. The results underscore the effectiveness of our proposed method in comparison to other attacks, demonstrating its capability to disrupt the caption generation process by inducing semantic errors.}
\label{tab:caption}

\end{table}

%% file: table/visual_grounding.tex
\begin{table}[]
\centering
\begin{tabular}{@{}l|ccc@{}}
\toprule
\textbf{Attack} & Val           & TestA         & TestB         \\ \midrule
Baseline        & 58.4          & 65.9          & 46.2          \\
SGA             & 56.5          & 63.7          & 45.4          \\
OT-Attack (Ours)            & \textbf{56.3} & \textbf{63.5} & \textbf{45.0} \\ \bottomrule
\end{tabular}
\caption{Performance on Visual Grounding Task Across RefCOCO+ Subsets. This table evaluates the success of different adversarial attacks across the validation (Val), TestA, and TestB subsets of the RefCOCO+ dataset. The TCL model served as the source for generating adversarial attacks, while the ALBEF model was the target for evaluating their effectiveness. Lower scores indicate a more successful adversarial attack, highlighting the inverse relationship between the evaluation metrics and the attack's performance. The results demonstrate the comparative effectiveness of our proposed methodology in reducing the ALBEF model's accuracy in the visual grounding task, as evidenced by the decreased performance across all subsets.}
\label{tab:visual_grounding}
\end{table}

%% file: section/Conclusion.tex
\section{Conclusion}
In this paper, we focus on improving the adversarial transferability of vision-language pre-training models. In detail, recent works have found that using inter-modality interaction and data augmentation can significantly enhance the transferability of adversarial examples for vision-language pre-training models. Unfortunately, previous works ignore the optimal alignment problem between data-augmented image-text pairs, which leads the generated adversarial examples overfit to the source model and achieves limited adversarial transferability improvement. To address the issue, we propose an Optimal Transport-based Adversarial Attack, \emph{dubbed} OT-Attack. The proposed OT-Attack formulates the features of image and text sets as two distinct distributions, leveraging optimal transport theory to identify the most efficient mapping between them. It utilizes their mutual similarity as the cost matrix. The derived optimal mapping guides the generation of adversarial examples, effectively mitigating overfitting issues and improving adversarial transferability. Extensive experiments across diverse network architectures and datasets in image-text matching tasks demonstrate the superior performance of the proposed OT-Attack compared to existing methods in terms of adversarial transferability. Significantly, our results also show that OT-Attack is also effective in cross-task attacks, including image captioning and visual grounding, and poses a considerable challenge to commercial models such as GPT-4 and Bing Chat, highlighting the evolving landscape of adversarial threats in advanced AI applications. 
This underscores the need for robust defenses against sophisticated attacks.